\documentclass[preprint,merge, final]{elsarticle}

\usepackage{amsmath}
\usepackage{bm}
\usepackage{graphicx}
\usepackage{siunitx}

\usepackage{tikz}
\usetikzlibrary{fit}
\usepackage{xifthen}
\usepackage[notcite, notref, color]{showkeys}
\usepackage[table]{xcolor}
\usepackage{listings}
\newcommand{\logit}{\operatorname{logit}}

\newcommand{\sless}[1]{s_{<#1}}
\newcommand{\sleq}[1]{s_{\leq#1}}
\newcommand{\sgreat}[1]{s_{>#1}}
\newcommand{\sgeq}[1]{s_{\geq{#1}}}
\newcommand{\cond}[2]{#1(s_{#2}|\sless{#2})}
\newcommand{\s}[1]{s_{#1}}
\newcommand{\bs}{\bm{s}}
\newcommand{\ps}{p(\bs)}
\newcommand{\pt}{\tilde{p}}
\newcommand{\Ps}{P(\bs)}
\newcommand{\qs}{q(\bs)}
\newcommand{\lnk}[2]{\s{#1}\s{#2}}
\newcommand{\ptn}[1]{\pt^{(#1)}}

\newcommand{\avg}[1]{\left\langle{#1}\right\rangle}

\newcommand{\eval}[1]{\the\numexpr#1\relax}

\begin{document}
\begin{frontmatter}
\title{Variational Autoregressive Networks with probability priors.}
\author[1]{Piotr Białas\corref{cor}}
\ead{piotr.bialas@uj.edu.pl}
\author[2]{Piotr Korcyl}
\ead{piotr.korcyl@uj.edu.pl}
\author[2]{Tomasz Stebel}
\ead{tomasz.stebel@uj.edu.pl}
\author[1,3]{Dawid Zapolski}
\ead{dawid.zapolski@doctoral.uj.edu.pl}

\cortext[cor]{Corresponding author.}

\affiliation[1]{organisation={Institute of Applied Computer Science, Jagiellonian~University}, 
addressline={ul.~Łojasiewicza~11}, 
postcode={30-348}, city={Kraków}, country={Poland}}
\affiliation[2]{
organisation={Institute of Theoretical Physics, Jagiellonian~University},
addressline={ul.~Łojasiewicza~11},
postcode={30-348}, city={Kraków}, country={Poland}
}
\affiliation[3]{organisation={Doctoral School of Exact and Natural Sciences, Jagiellonian~University}, addressline={ul.~Łojasiewicza~11}, postcode={30-348}, city={Kraków}, country={Poland}}

\begin{abstract}
Monte Carlo methods are essential across diverse scientific fields, yet their efficiency is frequently hampered by critical slowing down—a sharp increase in autocorrelation times near phase transitions. Although deep learning approaches, such as neural-network-based samplers, have been proposed to alleviate this issue, they face another serious problem: the difficulty of training the models. 
This difficulty partially stems from the overly general nature of original machine-learning architectures, which often ignore underlying physical symmetries and force networks to relearn them from scratch. In this paper, we demonstrate that incorporating physical priors into the model significantly enhances performance. Building upon existing strategies that integrate spin-spin interactions, we propose a framework that utilizes a prior probability distribution as a starting point for training. Our results for the Ising model, as well as for the Edwards-Anderson spin glass model, suggest that moving away from "blank slate" models in favor of physics-informed priors reduces the training burden and facilitates the simulation of larger system sizes in discrete spin models.    
\end{abstract}

\end{frontmatter}

\section{Introduction}

Monte-Carlo methods play a very important role in many areas, ranging from social sciences, through quantitative finance, to physics. The generation of samples is often done by some variant of the Metropolis-Hastings algorithm, which resorts to the construction of a Markov chain of consecutive configurations \cite{metropolis, *Hastings}. However, by the nature of this construction, which builds the new configuration based on the previous one, the configurations are correlated. Thus, the efficiency of the simulation depends not only on how fast the configurations can be generated but also on the number of configurations that we have to generate before they are independent. This is quantified by the {\em autocorrelation} time. Unfortunately, in many cases, this autocorrelation time grows in the interesting region, i.e.~near the phase transition. This undesired phenomenon is called {\em critical slowing down} and is the bane of lattice QCD.   

With the advent of deep learning, there have been renewed efforts to apply machine-learning techniques to alleviate this problem. In a seminal work \cite{VANPRL}, the authors proposed using neural networks to generate samples from a target distribution for discrete spin systems\footnote{For continuous models, like e.g.~$\phi^4$ theory, an approach based on normalizing flows was proposed in \cite{PhysRevD.100.034515}; however, in this contribution, we restrict ourselves solely to discrete spin systems.}.   Although this approach significantly reduces autocorrelation times \cite{Bialas:2021bei}, the system sizes that could be simulated remain limited compared to those accessible with specialized Monte-Carlo methods. Although the generation of samples is relatively fast and the samples are not correlated, the training of the neural networks has emerged as the main computational bottleneck.

We believe that the difficulty of training comes, at least in part, from the very general nature of the original proposal in \cite{VANPRL}. If we restrict our consideration to two-state spin systems, then the only input to the training algorithm is the energy of a configuration. Moreover, the factorization of the Boltzmann probability distribution into a product of conditional probabilities required by this approach destroys the existing symmetries. Those symmetries must be relearned by the networks, which increases the difficulty of the training.  Many of the improvements that allowed simulation of larger system sizes were due to the incorporation of at least some part of physics \cite{Biazzo2023, Biazzo2024, Bialas:2022qbs,singha2025, 2025MLS&T6b5029D}. 

In this contribution, instead of starting from a blank slate, we propose to add a {\em prior} probability distribution as a starting point for the training. This can be thought of as an extension of the approach presented in \cite{Biazzo2023, Biazzo2024} and \cite{2025MLS&T6b5029D}, where some of the spin-spin interactions were incorporated into the model, leading to faster training times. 

This work is organized as follows. In the next chapter, we introduce autoregressive neural generators. Next, we review some of the methods that aim to improve a simple approach presented in \cite{VANPRL}. In the following chapter, we show how to augment this approach when we have some approximation of the desired probability distribution. The rest of the work presents the results for the ferromagnetic Ising model and the Edwards-Anderson spin glass model \cite{Edwards_1975}.


\section{Autoregressive neural generators}

Given some set of $N$ spins, 
\begin{equation}
    \bm{s}=(s_0,s_1,\ldots,s_{N-1})
\end{equation}
we will be concerned with the Boltzmann distribution\footnote{Please note that most distributions can be put in this form. } 
\begin{equation}
\label{eq-boltzman}
    p(\bm{s}) = \frac{1}{Z(\beta)}e^{-\beta E(\bm{s})}
\end{equation}
where $\beta=\left(k_B T\right)^{-1}$ is the inverse temperature, $E(\bs)$ is the energy of configuration $\bs$ and 
\begin{equation}
Z(\beta) = \sum_{\bm{s}} e^{-\beta E(\bm{s})}   
\end{equation}
is the so called {\em partition function}. In practice, we seldom know the partition function $Z(\beta)$, so we only have access to unnormalized probability 
\begin{equation}
\label{eq:P}
P(\bm{s}) =  e^{-\beta E(\bm{s})}.
\end{equation}

The idea of {\em neural generators} is to train a model to approximate the probability distribution $\ps$ by some other probability distribution $q(\bm{s})$. For discrete systems, we represent the probability $q(\bm{s})$ as a product of conditional probabilities 
\begin{equation}
\label{eq:q-factorisation}
q(s_0,s_1,\ldots,s_{N-1}) = q(s_0)\prod_{i=1}^{N-1}q(s_i|s_{<i})
\end{equation}
where
$N$ is the number of spins and 
\begin{equation}
    s_{<i}\equiv(s_0,s_1,\ldots,s_{i-1})
\end{equation}
or in general 
\begin{equation}
    s_{\text{op } i}\equiv\{s_j : \text{j op i = true}\}.
\end{equation}
where \texttt{op} is any operator used to compare indices $i$ and $j$. 
In this way, we can use some ML model for $\cond{q}{i}$. 
Any chosen model must have the {\em autoregressive} property: the output $i$ can only depend on the inputs $j<i$. 
In the simplest example of a system of spins that can take only two values of $\pm1$ we need a model with $N$ inputs and $N$ outputs. Each input corresponds to one spin and each output to the conditional probability $q(s_i=1|\sless{i})$. In the following, we will use the shorthand
\begin{equation}
q_i(\sless{i})\equiv q(s_i=1|\sless{i}).     
\end{equation}
and similarly for $\cond{p}{i}$. 
The authors of \cite{VANPRL} call such models based on neural networks Variational Autoregressive Networks (VAN). 
Given that, we can sample the configurations $\bm{s}$ from $q(\bm{s})$ using ancestral sampling. Once we know the values of the spins $s_1,s_2,\ldots,s_{i-1}$, we sample the spin $s_i$ from the distribution $q(s_i|\sless{i})$ and so on.

The model is trained by minimizing the quantity that, with some abuse of notation (as it differs from free energy by a factor of $\beta^{-1}$), we will call $F_q$
\begin{equation}
\label{eq: F_q definition}
F_q =\sum_{\bm{s}}q(\bm{s})\log \frac{q(\bm{s})}{e^{-\beta E(\bm{s})}} =     
\sum_{\bm{s}}q(\bm{s})\log \frac{q(\bm{s})}{p(\bm{s})} -\log Z.
\end{equation}
Defining $F=-\log Z$ we obtain
\begin{equation}
\label{eq:F_q}
F_q = D_{KL}(q||p) + F.    
\end{equation}
So, minimizing $F_q$ is equivalent to minimizing the (reverse) Kullback-Leibler divergence\footnote{$F_q$ is related to the so-called evidence lower bound (ELBO): ELBO=$-F_q$, see for example Ref.~\cite{2024arXiv240607423B}.} $D_{KL}(q||p)$.  
This can be done using the REINFORCE algorithm as described in \cite{VANPRL}. The $F_q$ can be estimated by sampling from $\qs$:
\begin{equation}
\label{eq:F-q-estimator}
F_q\approx \frac{1}{M}\sum_{i=1}^M \left(\log q(\bs^i) +\beta E(\bs^i)\right),\qquad \bs^i\sim q(\bs^i).    
\end{equation}

The choice of architecture that would embody the probabilities $q(s_i|s_{<i})$, while crucial, is arbitrary. In the original reference \cite{VANPRL}, the authors considered both fully connected \cite{2015arXiv150203509G} and convolutional networks with masks that ensure the autoregressive property. 

In practice, the distribution $q(\bm{s})$ will never be equal to $p(\bm{s})$, but if it is close enough, we can still use these samples to produce samples from $p(\bm{s})$ \cite{PhysRevD.100.034515, PhysRevE.101.023304}.  One way is to use the samples generated from $q(\bm{s})$ as the proposal in the Metropolis-Hastings algorithm, accepting them with probability
\begin{equation}
\min\left(1,
\frac{p(\bm{s}^{i+1})}{p(\bm{s}^{i}) }\frac{q(\bm{s}^{i})}{q(\bm{s}^{i+1}) }
\right)    =  
\min\left(1,
\frac{P(\bm{s}^{i+1})}{P(\bm{s}^{i}) }\frac{q(\bm{s}^{i})}{q(\bm{s}^{i+1}) }
\right). 
\end{equation}
This step introduces correlations, but for $q(\bm{s})$ close to $\ps$ those correlations will be substantially smaller \cite{Bialas:2021bei}. 

Another way is to use {\em importance sampling}. Given some observable $O$, we can calculate its average as
\begin{equation}
\begin{split}
\avg{O}_{p} & \equiv \sum_{\bs}\ps O(\bs)   =    \sum_{\bs}\qs\frac{\ps}{\qs} O(\bs) \\
&\approx \frac{1}{M}\sum_{i=1}^M w(\bs) O(\bs^i),\quad \bs^i\sim q(\bs^i)
\end{split}
\end{equation}
where $\bs^i\sim q(\bs^i)$ denotes that the samples $\bs^i$ are sampled with probability $q(\bs^i)$. 
The $w(\bs)$ are the so called importance weights
\begin{equation}
w(\bs)=\frac{\ps}{\qs}.     
\end{equation}
As we do not know $\ps$, but only $P(\bs)$, the final approximation of  $\avg{O}_{p}$ is
\begin{equation}
\avg{O}_{p} \approx\frac{ \frac{1}{M}\sum_{i=1}^M \tilde{w}(\bs^i) O(\bs^i) }{
\frac{1}{M}\sum_{i=1}^M \tilde{w}(\bs^i) 
}   
\end{equation}
where
\begin{equation}
\tilde{w}(\bs)=\frac{\Ps}{\qs}=Zw(\bs).     
\end{equation}

An interesting corollary is that
\begin{equation}
\label{eq:Z-nis}
\avg{\tilde{w}(\bs)}_{\qs} = Z \avg{w(\bs)}_{\qs} = Z \sum_{\bs}\qs \frac{\ps}{\qs} = Z   
\end{equation}
which makes
\begin{equation}
\label{eq:Z_nis}
Z_{nis}=\frac{1}{N}\sum_{i=1}^N \frac{e^{-\beta E(\bs)}}{q(\bs)},\quad \bs \sim q(\bs)    
\end{equation}
an unbiased estimator of $Z$ \cite{PhysRevE.101.023304}. This makes this approach attractive even for relatively small systems, as it allows for a precise estimation of $Z$ and related quantities such as mutual information and entanglement entropy \cite{PhysRevE.110.044116, PhysRevE.108.044140, PhysRevLett.134.151601} and even the whole reduced density matrix \cite{bialas2025estimationreduceddensitymatrix}. Those quantities are usually hard to estimate using Monte-Carlo methods.

\section{Physics aware neural generators}

As already pointed out in the Introduction, the approach outlined above is very general. This can be an advantage, as we only need to provide an energy function definition to train an arbitrary model. However, when simulating more restricted models, e.g. with only nearest neighbor interaction and some symmetries, the model has to unnecessarily relearn those properties. 

\begin{figure}
    \begin{center}
\begin{tikzpicture}[scale=0.9,thick , x=1cm, y=1cm, every node/.style={circle, draw, minimum size=7mm, inner sep=0pt, scale=1.0, font=\small}]
  \def\rows{7}
  \def\cols{7}

  \foreach \i in {0,...,\rows} {
    \foreach \j in {0,...,\cols} {
      \newcommand{\n}{\eval{8*\i+\j}}
      \newcommand{\offset}{\eval{\j-3}}
    \newcommand{\coffset}{\eval{\i-1}}
      \newcommand{\lab}{\n}

      \ifthenelse{\n=27}{\node [double, draw=red] (s\n) at (\j, 7-\i) {$s_{\lab}$};}{
        \ifthenelse{\n<27}{\node (s\n) at (\j, 7-\i) {$s_{\lab}$};}{\node [dashed] (s\n) at (\j, 7-\i) {$s_{\lab}$};}
      }
    }
  }

\draw [black, very thick] (s19)--(s27)--(s26);

\draw [blue, very thick]  (s20)--(s28)--(s27); 

\draw [green, very thick] (s21)--(s29)--(s28);
\draw [green, very thick] (s26)--(s34)--(s35)--(s27); 

\draw [magenta, very thick] (s25)--(s33)--(s34);
\draw [magenta, very thick] (s28)--(s36)--(s35); 
\draw [magenta, very thick] (s22)--(s30)--(s29);

\foreach \i in {0,...,\eval{\cols+3}} {
    \node [dotted] (b\i) at(\i,8) {$0$};
}

\foreach \i in {0,...,\eval{\rows+1}} {
    \node [dotted] (l0\i) at (-1,\i) {$0$};
    \node [dotted] (l1\i) at (-2,\i) {$0$};
}

\foreach \i in {0,...,\rows} {
    \node [dotted] (r\i) at (8,\i) {$s_{\eval{8*(7-\i)}}$};
}
\foreach \i in {0,...,\rows} {
    \node [dotted] (rr\i) at (9,\i) {$s_{\eval{8*(7-\i)+1}}$};
}
\foreach \i in {0,...,\rows} {
    \node [dotted] (rrr\i) at (10,\i) {$s_{\eval{8*(7-\i)+2}}$};
}
\definecolor{lightgray}{rgb}{0.7,0.7,0.7}
\node[rectangle, rounded corners=2mm, fit = (s0) (s63), draw=gray, inner sep= 1mm ] {}; 
\node[rectangle, rounded corners=2mm, fit = (s17) (s30), draw=lightgray, inner sep= 1mm ] {};

\end{tikzpicture}
\end{center}
    \caption{Calculating the $\cond{p}{27}$. Solid black lines indicate the spins that are already fixed. The dashed lines denote the spins that are not yet fixed and should be summed over to obtain the true conditional probability $p_{27}(\sless{27}
    )$. The dotted lines indicate padding spins that are required for efficient  $t^4$ order calculations. The inner contour represents the convolution kernel for $t^4$ order approximation. Other approximations require smaller kernels and respectively smaller padding. This approach does not take into account the periodic boundary conditions in the vertical direction.} 
    \label{fig:p-cond-27}
\end{figure}
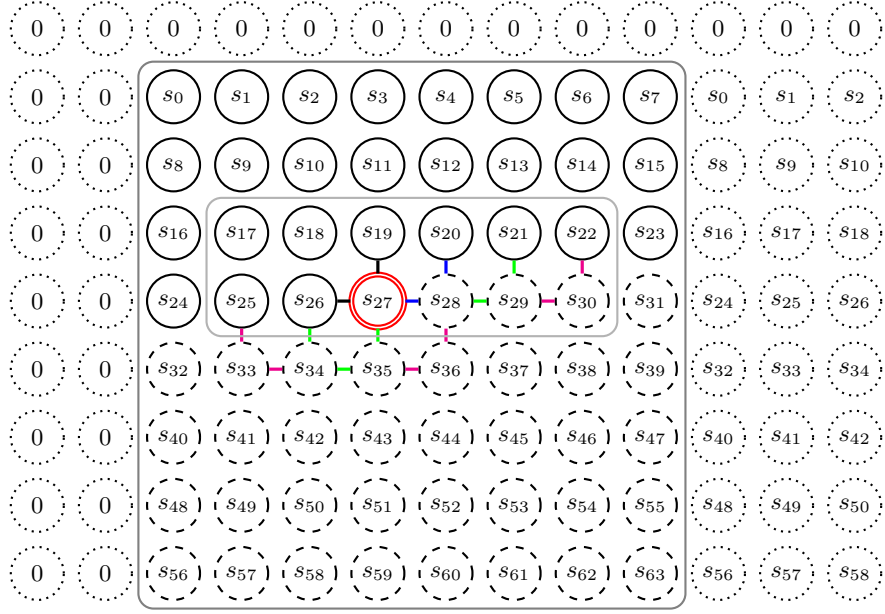

For example, the probability $q(s_i|\sless{i})$ depends on all spins $s_j$ with $j<i$. However, for models with nearest neighbor interactions this dependence can be significantly reduced. For a two dimensional system, assuming that we order spins row by row, only two rows contribute due to the Markov property.
For example, in Figure~\ref{fig:p-cond-27} $q(s_{27}|\sless{27})$ 
depends only on the spins $s_0$ to $s_7$ and  $s_{19}$ to $s_{26}$. This restricted dependency is not taken into account by simple VANs as proposed in \cite{VANPRL}. Hierarchical autoregressive networks introduced in \cite{Bialas:2022qbs} alleviate this problem to some extent, resulting in superior training results. This approach also restores some of the translational symmetry broken by factorization \eqref{eq:q-factorisation}. 
Other contributions also exploit the multilevel renormalization group structure of nearest neighbor statistical models \cite{singha2025}. 

Another issue is the symmetries, which play a very important role with regard to the properties of the system, especially an internal symmetry such as $Z_2$. On the level of the conditional probabilities,  the $Z_2$ symmetry requires
\begin{equation}
    q_i(-\sless{i})=1-q_i(\sless{i})
\end{equation}
which is not enforced by the \verb|LeakyReLU| activation functions typically used in VANs. This symmetry can be partially enforced by averaging probabilities in symmetric configurations \cite{VANPRL,n-n-renormalization-group}. This approach can be extended to other symmetries, such as translational symmetry, which is also broken by factorization
 \eqref{eq:q-factorisation} \cite{Bialas:2021bei,Bialas:2022bdl}. Again, this leads to better training results. 

The authors of \cite{Biazzo2023, Biazzo2024} aim to include in the model some of the interactions of the Hamiltonian $H(\bs)$. In this contribution, we take the same route, but our approach is less rigorous and more general, as explained in the next section.


\section{Approximate conditional probabilities}
\label{sec:approx}

Probability distribution $\ps$ can be factorized in the same way as $\qs$  
\begin{equation}
\label{eq:p-factorisation}
p(s_0,s_1,\ldots,s_{N-1}) = p(s_0)\prod_{i=1}^{N-1}p(s_i|s_{<i}). 
\end{equation}
Our idea is to start with a approximation for each conditional probability 
\begin{equation}
p(s_i|\sless{i})\approx \cond{\pt}{i}
\end{equation}
and use this expression as a starting point for training the neural network 
\begin{equation}
\label{eq: q_cond_prob}
q(s_i|s_{<i}) = \frac{1}{1+e^{\displaystyle-h^{n-1}_i -\logit(\cond{\tilde{p}}{i})}},
\end{equation}
where $h^{n-1}_i$ is the output of the last layer of the neural network modeling $\qs$ and $\logit$
is the inverse of the logistic $\sigma$ function (see Figure~\ref{fig:implementation})
\begin{equation}
\sigma(x)\equiv\frac{1}{1+e^{-x}}, \qquad   
    \logit(y)\equiv \sigma^{-1}(y) =\log\frac{y}{1-y}.
\end{equation}
In this way, the model has only to learn the difference between distributions $\tilde{p}(\bs)$ and $\ps$. If the approximation is good, we can expect reduced training times as well as better final quality of the model.  

By definition
\begin{equation}
    p(s_i|\sless{i})\equiv\frac{p(\sleq{i})}{p(\sless{i})}=
    \frac{\sum_{\sgreat{i}}e^{-\beta E(\bs)}}{\sum_{\sgeq{i}}e^{-\beta E(\bs)}}.
\end{equation}    
The denominator is just a normalization factor so that 
\begin{equation}
    \sum_{s_i=\pm1}\cond{p}{i} = 1 .
\end{equation}
Combining this with the definition of logit, we obtain that
\begin{equation}
\label{eg-logit-p-i}
    \logit{p_i(\sless{i})} = \log
    \frac{\left.\sum_{\sgreat{i}}e^{-\beta E(\bm{s})}\right|_{s_i=1}}
    {\left.\sum_{\sgreat{i}}e^{-\beta E(\bm{s})}\right|_{s_i=-1}}.
\end{equation}
The sums in the above expression can be further decomposed as 
\begin{equation}
\begin{split}
\sum_{\sgreat{i}}e^{-\beta E(\bm{s})} = e^{-\beta I(\sless{i})}
\sum_{\sgreat{i}}e^{-\beta E(\{\sgeq{i}\})}
\end{split}    
\end{equation}
where $I(\{s\})$ denotes the {\em internal} energy of the set of spins $\{s\}$, {\em i.e.} , the energy of all the bounds directly connecting two spins in $\{s\}$. The $E(\{s\})$ denotes the energy of all the bounds originating from any spin in the set $\{s\}$.  The first factor does not depend on $s_i$, so it will cancel out in \eqref{eg-logit-p-i}, leaving 
\begin{equation}
\label{eq:logit-p-i-2}
    \logit p_i(\sless{i}) = \log 
    \frac{\left.\sum_{\sgreat{i}}e^{-\beta E(\sgeq{i})}\right|_{s_i=1}}
    {\left.\sum_{\sgreat{i}}e^{-\beta E(\sgeq{i})}\right|_{s_i=-1}}.
\end{equation}
The formula \eqref{eq:logit-p-i-2} will be our starting point for approximations.

In this contribution, we consider nearest neighbor spin glass systems with energy given by 
\begin{equation}
\label{eq:energy}
E(\bs)=-\sum_{<i,j>}J_{i,j}s_i s_j    
\end{equation}
where $J_{i,j}$ are some arbitrary {\em link variables}. For $J_{i,j}=1$ this reduces to the ferromagnetic Ising model as a special case.    

We will also use another representation for the Boltzmann factor \eqref{eq:P}. As spins can take only values $\pm1$, we have 
\begin{equation}
e^{\beta J_{ij} s_i s_j} = 
\begin{cases}
e^{\beta J_{ij}} & s_i s_j = 1\\
e^{-\beta J_{ij}} & s_i s_j = -1
\end{cases}
\end{equation}
which allows us to write
\begin{equation}
\label{eq:weak}
\begin{split}
e^{\beta J_{ij} s_i s_j } &= \frac{1}{2}(1+s_i s_j)e^{\beta J_{ij}} + \frac{1}{2}(1-s_i s_j)e^{-\beta J_{ij}}\\
&=\frac{1}{2}(e^{\beta J_{ij}}+e^{-\beta J_{ij}}) +s_i s_j \frac{1}{2}(e^{\beta J_{ij}}-e^{-\beta J_{ij}})\\
&=\cosh(\beta J_{ij}) +s_i s_j \sinh(\beta J_{ij})\\
&= \cosh(\beta J_{ij})\left(1+s_i s_j\tanh(\beta J_{ij})\right).
\end{split}
\end{equation}
When $J_{i,j}=1$ this reduces to 
\begin{equation}
\label{eq:weak-ising}
\cosh(\beta)\left(1+s_i s_j\tanh(\beta)\right).
\end{equation}

 We will approximate $p(s_i|\sless{i})$ taking only a subset of spins from $\sless{i}$ and sum only over a subset of spins from $\sgreat{i}$, neglecting the dependence  on any other spins.  As an example, we will use the probability $p_{27}(\sless{27})$ (see Figure~\ref{fig:p-cond-27}). In this section, we will concentrate on the Ising model setting $J_{i,j}=1$. The formulas for the general spin glass case will be given in Section~\ref{sec:ea}.

The simplest approximation would be to consider only the nearest neighbor interactions with spins $s_{19}$ and $s_{26}$ (black links in Figure~\ref{fig:p-cond-27}) which gives the expression
\begin{equation}
\label{eq:ising-logit-1}
\begin{split}
\logit\left( \pt^{(1)}_{27}(\sless{27})\right) = 
\log\frac{e^{\beta (s_{19}+s_{26})}}
{e^{-\beta(s_{19}+s_{26})}} = 2\beta (s_{19}+s_{26}).
\end{split}
\end{equation}
This corresponds to the approach taken in \cite{Biazzo2023, Biazzo2024}. Please note, however, that those references use different neural network architectures, so the results are not directly comparable.  

In the next approximation, we add the dependence on spin $\s{20}$ and sum over spin $\s{28}$. Using the representation \eqref{eq:weak-ising}, we obtain 
\begin{equation}
\begin{split}
\sum_{s_{28}}
e&^{\beta(s_{27}s_{28}+s_{28}s_{20})}\\
&=\left(\cosh\beta\right)^2\sum_{s_{28}}(1+ s_{27}\s{28}t)(1+\s{28}\s{20}t) = \left(\cosh\beta\right)^2(1+s_{27}\s{20}t^2),  
\end{split}
\end{equation}
where $t=\tanh\beta$ and 
\begin{equation}
\label{eq:ising-logit-2}
\begin{split}
\logit \pt^{(2)}_{27}(\s{19},s_{20},\s{26})&=2\beta (s_{19}+s_{26})+\log\left(
\frac{1+\s{20}t^2}{1-\s{20}t^2}\right)\\
&\approx 2\beta (s_{19}+s_{26}) + 2 \s{20}t^2.
\end{split}
\end{equation}

For the order $t^3$ calculations, we add the dependence on spin $\s{21}$ and additionally sum over the spins $\s{29}$, $\s{34}$ and $\s{35}$. This requires the calculation of the sum
\begin{equation}
\begin{split}
\sum_{\s{28},\s{29},\s{34},\s{35}}& e^{\beta(\lnk{26}{34}+\lnk{34}{35}+\lnk{35}{27}+\lnk{27}{28}+\lnk{28}{20}+\lnk{28}{29}+\lnk{29}{21})} \\
&= \sum_{\s{34},\s{35}} e^{\beta(\lnk{26}{34}+\lnk{34}{35}+\lnk{35}{27})}
\sum_{\s{28},\s{29}}
e^{\beta(\lnk{28}{20}+\lnk{28}{29}+\lnk{29}{21})}.
\end{split}    
\end{equation}
Going through the same steps as before, we obtain 
\begin{equation}
\label{eq:ising-logit-3}
\begin{split}
\logit\pt^{(3)}_{27}(\s{19},s_{20},\s{21},\s{26} )
= 2\beta (s_{19}+s_{26}) + 2 \s{20}t^2 +2(\s{21}+\s{26})t^3.
\end{split}
\end{equation}
Finally, we present the results for the approximation $t^4$, where we add dependence on spins $\s{22}$ and $\s{25}$ and additionally sum over spins $\s{30},\s{33}$ and $\s{36}$
\begin{equation}
\label{eq:ising-logit-4}
\begin{split}
\logit\ptn{4}_{27}&(\s{19},s_{20},\s{21},\s{22},\s{25},\s{26})\\
&= 2\beta (s_{19}+s_{26}) + 2 \s{20}t^2 +2(\s{21}+\s{26})t^3 \\
&\phantom{=2\beta}+2(\s{20}+\s{22}+\s{25})t^4. 
\end{split}
\end{equation}
Those calculations were performed using the \texttt{Mathematica} software. Formulae \eqref{eq:ising-logit-1}, \eqref{eq:ising-logit-2}, \eqref{eq:ising-logit-3}, \eqref{eq:ising-logit-4} can be easily translated to match any other spin, but they do not take into account the periodic boundary conditions in the vertical direction. That means that the formulas for the spins in the last rows are, strictly speaking, not correct. However, as those formulas are approximations anyway, we just assume that they are steps in a good direction and favor the ease of implementation over correctness. 

One way of testing the quality of these approximations is to calculate the $F_q$ for each of them and compare it with the true value that can be obtained using formulas from \cite{Ising-model-on-finite-lattice}. Those numbers are presented in Table~\ref{tab:F-q}. $F_q$ was estimated using the formulas \eqref{eq:F-q-estimator} and \eqref{eq: q_cond_prob} setting $h^{n-1}_i=0$. It is worth noting in Table~\ref{tab:F-q} that already $t^1$ provides a noticeable improvement with respect to an untrained neural network for all the temperatures considered. The value for $t^0$ was calculated analytically, assuming a uniform distribution $q(\bs)=2^{-N}$, where $N$ is the number of spins. This was done using another representation for $F_q$: 
\begin{equation}
F_q = \sum_{\bs} q(\bs) \log \qs    +\beta\sum_{\bs} \qs E(\bs) =\beta\avg{E(\bs)}_q -S_q, 
\end{equation}
where $S_q$ is the entropy of the distribution $\qs$. For a uniform distribution, this equals $N\log(2)$. If one colors the spins in a checkerboard fashion and flips every white spin, the energy of the configuration changes sign. As both flipped and non-flipped configurations are equally probable in this case, we obtain a zero average energy, leading to the final result, $F_q=-N\log(2)$, presented in Table~\ref{tab:F-q}. 

\sisetup{
  group-digits = false
}
\begin{table}
\small
    \centering
    \begin{tabular}{|r|S[table-format=4.5]|S[table-format=4.5]|S[table-format=4.5]|S[table-format=5.5]|}
    \hline\hline
    & \multicolumn{4}{c|}{$\beta$}\\\hline\hline
            &  {0.40} & {0.42} & {$\beta_c$} & {0.50}\\\hline
             $t^0$    &  \multicolumn{4}{c|}{-709.783} \\\hline        
    $t^1$   &  -874.490(7)  & -891.824(7)   &  -910.720(8)   & -970.19(1)\\
    $t^2$   &  -886.705(5)  & -906.461(6)   &  -928.152(6)   & -997.307(8)\\
    $t^3$   &  -891.059(4)  & -912.173(5)   &  -935.544(5)   & -1011.365(8) \\
    $t^4$   &  -892.393(4)  & -914.114(4)   &  -938.308(5)   & -1017.941(7) \\
     true   &  -900.478     & -924.4135     &  -952.648      & -1051.105\\
    \hline\hline  
    \end{tabular}
    \caption{Values of $F_q$ obtained with different conditional approximations for the Ising model on the $32\times 32$ lattice, $t^0$ denotes random distribution where each configuration has same probability which is a reasonable approximation for an untrained model.}
    \label{tab:F-q}
\end{table}


\section{Results -- Ising model}

\begin{figure}
    \centering
    \includegraphics[width=0.65\linewidth]{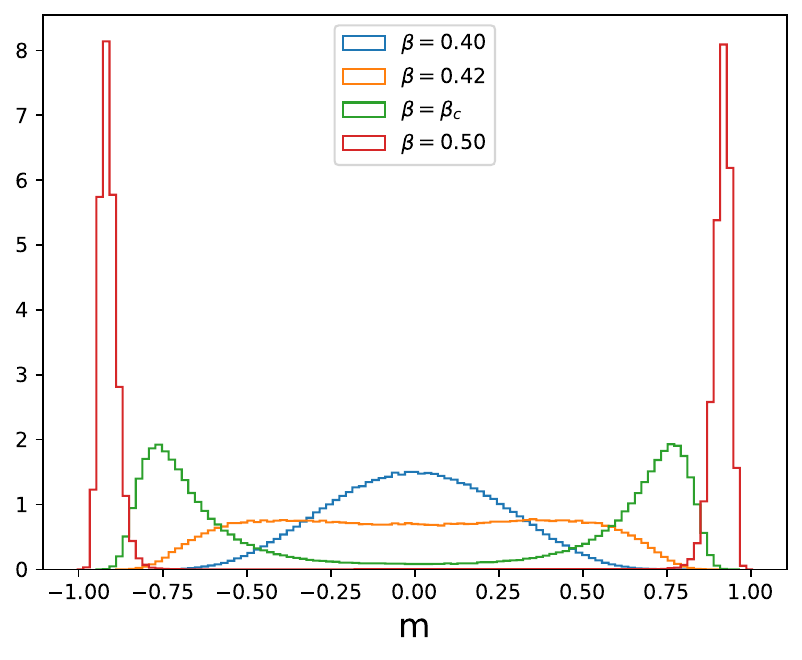}
\caption{Magnetization distribution for the values of $\beta$ considered in this reference. Obtained from Monte-Carlo simulations. }
    \label{fig:mag-hists}
\end{figure}

We started by testing our approach on a ferromagnetic Ising model on a $32\times 32$ lattice. In the absence of the magnetic field, this model has a $Z_2$ symmetry associated with the change of sign of each spin in the configuration, as such a transformation obviously leaves the energy \eqref{eq:energy} unchanged. In the thermodynamical limit, this model exhibits a continuous phase transition involving spontaneous breaking of this symmetry. In the high temperature symmetric phase,  the net magnetization 
\begin{equation}
\avg{m(\bs)}_{p(\bs)},\quad\text{where}\quad m(\bs) =\frac{1}{L^2}\sum_i s_i
\end{equation}
is zero. Below the critical temperature, in the broken phase,  a non-zero spontaneous magnetization appears. For the finite systems considered here, there is formally no phase transition and the average magnetization should always be zero. The transition, however, manifests itself through the presence of two peaks in the magnetization histogram (see Figure~\ref{fig:mag-hists}) and for low enough temperatures those peaks become totally separated. Monte Carlo simulation may become trapped in one of those regions with non-zero magnetization. The same problem can appear in neural samplers and is known as {\em mode collapse}.

We have chosen four different values of $\beta$: $(0.4, 0.42,\beta_c,0.5)$, where 
\begin{equation}
\beta_c = \frac{1}{2}\log(1+\sqrt{2}) \approx 0.4406868,     
\end{equation}
is the inverse critical temperature in the thermodynamic limit.
We have chosen them to cover the symmetric phase ($\beta=0.40$), the transition region ($\beta=0.42$) and the broken phase ($\beta=\beta_c$ and $\beta= 0.5$) (see Figure~\ref{fig:mag-hists}).   
We measure the quality of training using the estimated sample size (ESS) \cite{Kong, Liu1996},
\begin{equation}
\text{ESS} = \frac{\avg{\tilde{w}}_q^2}{\avg{\tilde{w}^2}_q}    
\end{equation}
and by comparing different estimates of $F$. We also track magnetization to see if the $Z_2$ symmetry is respected. 

For our investigations, we have chosen a very simple architecture consisting of two layers of fully connected neural networks with a leaky \texttt{ReLU} activation layer between them. As we wanted to assess only the influence of the probability priors, we did not add any enhancements related to symmetries or $\beta$ annealing.  All the models were implemented using the \verb|PyTorch| library. 
More details are presented in \ref{sec:implementation}. 

The summary of the results is presented in \ref{sec:results} in Tables~\ref{tab:ising-F} to \ref{tab:ising-EM-II}.
The Monte Carlo simulations were performed using the Wolff cluster algorithm \cite{Wolff}.  

\begin{figure}
    \centering
    \includegraphics[width=0.95\linewidth]{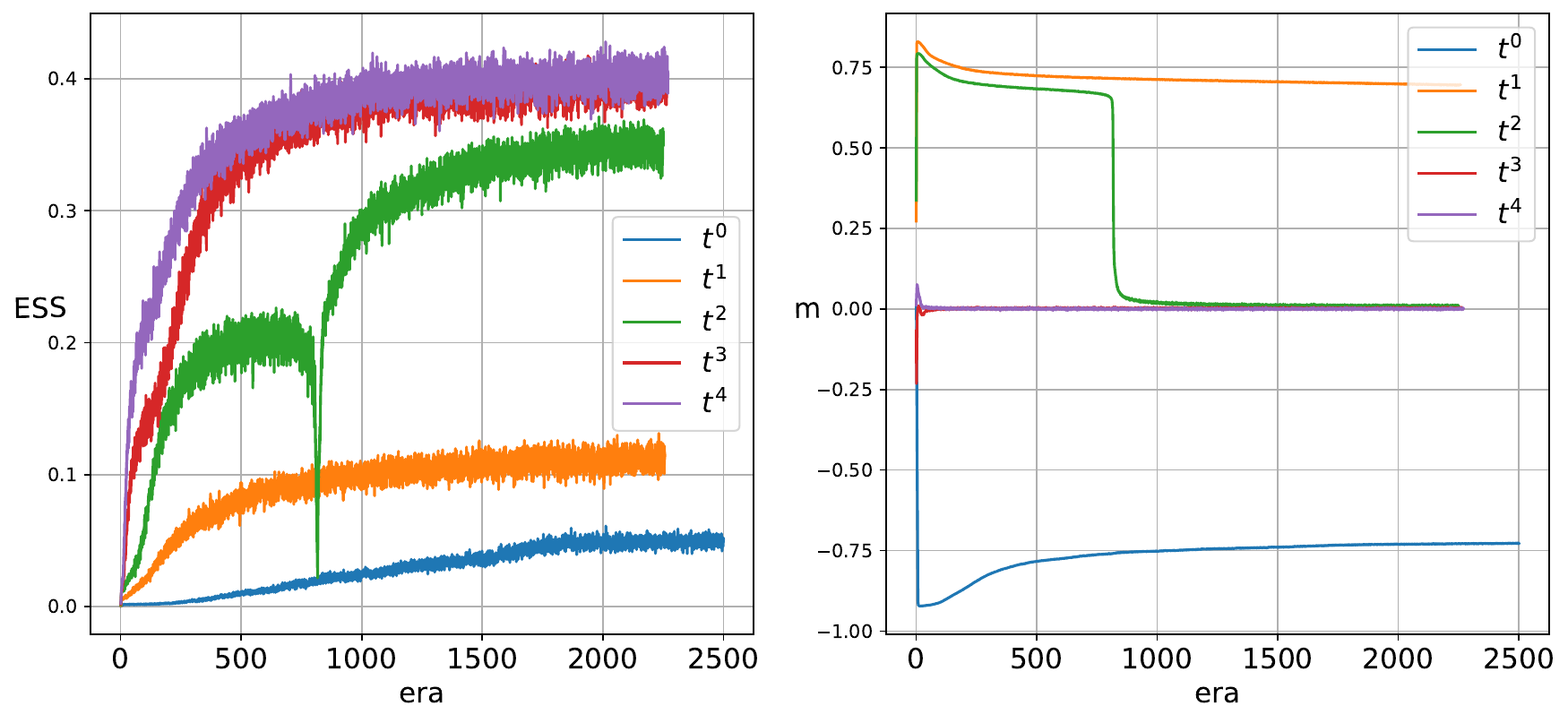}
    \caption{History of the training for Ising model on $32\time 32$ lattice at critical $\beta$. ESS (left) and $m$ (right). The horizontal axis marks the eras where an era is a 100 weights update using the batch of $4096$ samples. For clarity we present a moving average over $100$ weights updates.}
    \label{fig:ising-ess-32-bc}
\end{figure}
\begin{figure}
    \centering
    \includegraphics[width=0.5\linewidth]{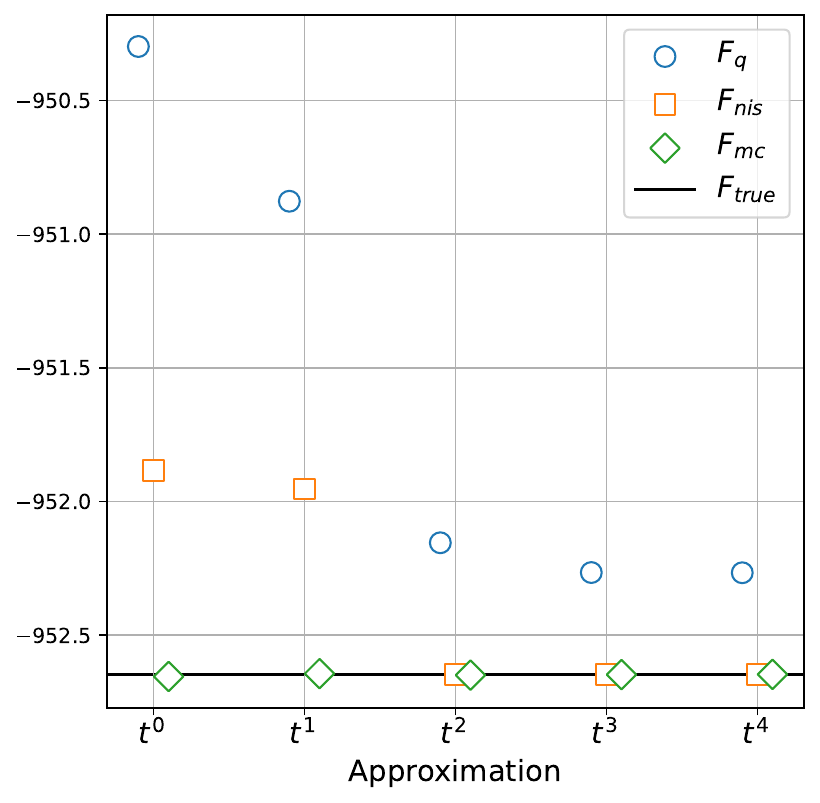}\includegraphics[width=0.5\linewidth]{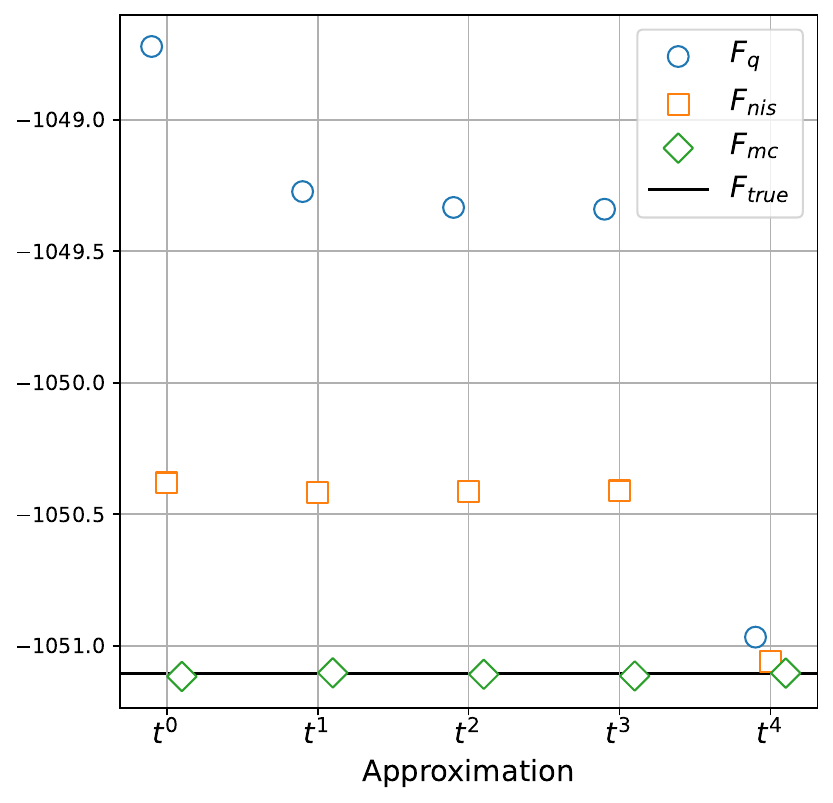}
    \caption{Various estimates of $F$ for the Ising model at critical $\beta$ (left) and $\beta=0.5$ (right). Uncertainties are much smaller than the points size.}
    \label{fig:F-bc}
\end{figure}

In Figure~\ref{fig:ising-ess-32-bc}  we present the history of the training for the model at critical temperature. We can see that the inclusion of approximate probabilities has a significant impact on the efficiency of the training. 
In particular, there is a notable change between the $t^1$ and $t^2$ approximations. The dip in ESS for the $t^2$ approximation corresponds to the restoration of the $Z_2$ symmetry, as can be seen in the plot showing magnetization in the right panel of Figure~\ref{fig:ising-ess-32-bc}. 
To test the quality of the trained models, we have generated $2^{20}$ configurations for each model and calculated  $F_q$ and  the estimate  derived from \eqref{eq:Z_nis}
\begin{equation}
F_{nis}=-\log Z_{nis}.    
\end{equation}
Using the configurations obtained from Monte-Carlo simulations, we have also calculated another estimator of $F$ \cite{DetectingModeColapse}
\begin{equation}
F_{mc} = -\log Z_{mc}.    
\end{equation}
where
\begin{equation}
\label{eq:Z-mc}
Z_{mc}^{-1}=\avg{\frac{q(\bs)}{e^{-\beta E(\bs)}}}_p 
\approx \frac{1}{M}\sum_{i=1}^M \frac{q(\bs^i)}{e^{-\beta E(\bs^i)}},\qquad \bs^i\sim p(\bs^i).
\end{equation}
It can be proven \cite{DetectingModeColapse} that
\begin{equation}
F_{nis}\ge F \ge F_{mc}. 
\end{equation} 
The discrepancy between those two estimators gives us an estimate of the quality of training and is an indicator of mode collapse. This can be quantified by introducing a parameter \cite{DetectingModeColapse}
\begin{equation}
\bar{w} \equiv \frac{Z_{nis}}{Z_{mc}} = e^{(F_{mc}-F_{nis})}. 
\end{equation}
The values of this parameter can be found in Table~\ref{tab:ising-F} in \ref{sec:results}.
The various estimators of $F$ for $\beta=\beta_c$ are compared in Figure~\ref{fig:F-bc} (left).  
We can see that starting with the approximation $t^2$, $F_{nis}$ and $F_{mc}$ coincide.

\begin{figure}
    \centering
    \includegraphics[width=\linewidth]{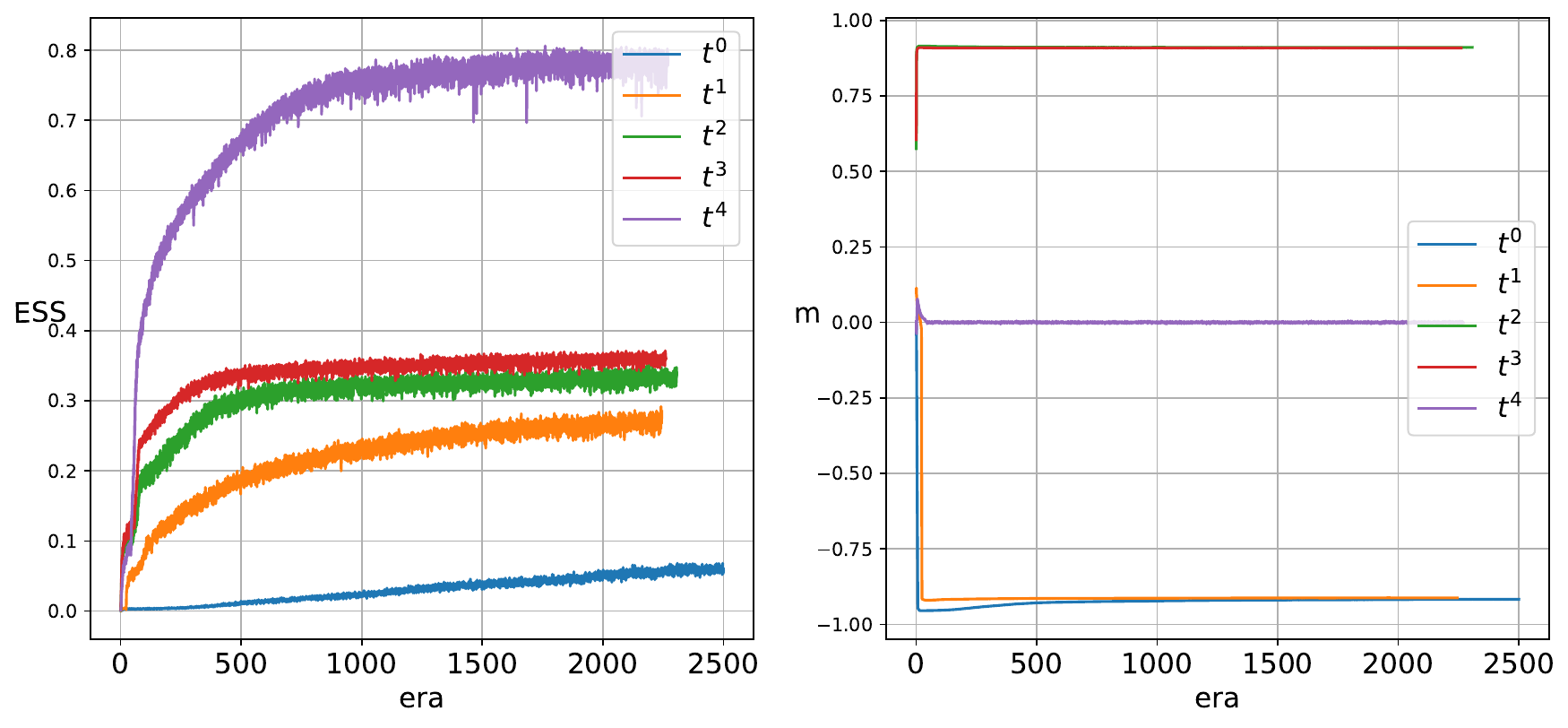}
    \caption{History of the training for the Ising model at $\beta=0.5$.}
    \label{fig:ising-ess-m-b0500}
\end{figure}

The expansion in $t$ that we use to derive the priors formally is divergent above the critical $\beta$. However, looking at the results for $\beta=0.5$ in Figure~\ref{fig:ising-ess-m-b0500}, we again see a significant improvement  with the increasing order of the approximation.
Actually, only the $t^4$ approximation was able to train properly and restore the $Z_2$ symmetry. Looking closely at the estimators of $F$ in Figure~\ref{fig:F-bc} (right), we observe that there remains a slight discrepancy between $F_{nis}$ and $F_{mc}$ resulting in $\bar{w}$ slightly below one and a small bias of $|M|$ obtained using NIS compared to Monte Carlo simulations (see Tables~\ref{tab:ising-F} and \ref{tab:ising-EM-II}).  

\section{Edwards-Anderson model}
\label{sec:ea}

In this section, we will consider the nearest neighbor spin glass model \cite{Edwards_1975} with energy given by \eqref{eq:energy},
where now the link variables take on the values $J=\pm1$ with equal probability. Spin glass models generally do not exhibit spontaneous symmetry breaking, and the phase transition, if present, is rather associated with {\em replica symmetry breaking} \cite{Edwards_1975,Parisi1983}. The two dimensional model that we simulate does not exhibit a phase transition at finite $\beta$ \cite{PhysRevB.22.288}, but it is increasingly difficult to simulate with growing $\beta$. 

By denoting
$t_{i,j} \equiv \tanh(\beta J_{i,j})$
and following the same derivation we have used in Section~\ref{sec:approx}, we obtain the approximations for a general spin glass model with nearest-neighbor interactions:
\begin{align}
\label{eq:ea-logit-1}
\logit\ptn{1}_{27}(\s{19},\s{26}) &= 2\beta(\s{19}J_{19,27}+ \s{26}J_{26,27}  ) \\
\label{eq:ea-logit-2}
\logit\ptn{2}_{27}(\s{19},\s{20},\s{26}) &= \logit\ptn{1}_{27}(\s{19},\s{26}) + 2\s{20}t_{20,28}t_{28,27}\\
\label{eq:ea-logit-3}
\logit\ptn{3}_{27}(\s{19},\s{20},\s{21},\s{26}) & = \logit\ptn{2}_{27}(\s{19},\s{20},\s{26}) + \nonumber\\
2 (\s{21}&t_{21,29}t_{29,28}t_{28,27} +\s{26}t_{26,34}t_{34,35}t_{35,27})\\
\label{eq:ea-logit-4}
\logit\ptn{4}_{27}(\s{19},s_{20},\s{21},\s{22},\s{25},\s{26})&=    \logit\ptn{3}_{27}(\s{19},\s{20},\s{21},\s{26})+\nonumber\\
2 \s{22} t_{22, 30}t_{30, 29} & t_{ 29, 28} t_{28, 27}  +
 2 \s{25} t_{25, 33}  t_{33, 34}  t_{34, 35} t_{35, 27} + \nonumber\\
 2 \s{20} t_{20, 28}t_{28, 36} & t_{36, 35} t_{35, 27} . 
\end{align}
In the case of link variables $J_{i,j}$ taking only the values $\pm 1$ 
\begin{equation}
t_{i,j}=\tanh(\beta J_{i,j}) = J_{i,j}\tanh(\beta) = J_{i,j} t.    
\end{equation}
and $t$ remains the expansion parameter. In Table~\ref{tab:ea-F-q} we present the values of $F_q$ obtained without any training, using only approximate probabilities. All the values are calculated for the same specific choice of link variables $J$. One should note that for $\beta=0.9$ the value of $F_q$ for the $t^4$ approximation is {\em higher} then the value for the $t^3$ approximation, indicating the possible divergence of the series in $t$. Those values were obtained in the same way as in the Ising model case, with the notable exception that the true value is unknown. 

\begin{table}
\definecolor{SoftGray}{RGB}{200,200,200}
\small
    \centering
    \begin{tabular}{|r|S[table-format=4.5]|S[table-format=5.5]|S[table-format=5.4]|}
    \hline\hline
    & \multicolumn{3}{c|}{$\beta$}\\\hline\hline
            &  {0.30} & {0.6} &  {0.9}\\\hline
    $t^0$    &  \multicolumn{3}{c|}{-709.783} \\\arrayrulecolor{SoftGray}\cline{2-4}
             \arrayrulecolor{black}\hline        
    $t^1$       &  -794.507(3)  &  -1007.76(1)   & -1282.50(2)  \\
    $t^2$       &  -797.584(3)  &  -1034.374(7)  & -1354.14(1)  \\
    $t^3$       &  -798.043(2)  &  -1043.466(6)  & -1379.10(1)   \\
    $t^4$       &  -798.087(2)  &  -1045.100(5)  & -1378.11(1)  \\
    \hline\hline  
    \end{tabular}
    \caption{Values of $F_q$ obtained with different conditional approximations for Edward-Anderson model on $32\times 32$ lattice for one specific choice of $J$, $t^0$ denotes random distribution where each configuration has same probability which is a reasonable approximation for an untrained model.}
    \label{tab:ea-F-q}
\end{table}

\section{Results - Edwards-Anderson model}

We have trained models for one particular set of link variables $J_{i,j}$ and three different  values of $\beta=(0.3,0.6,0.9)$. 
Because there are no analytical results for this model, the results were compared with Monte-Carlo simulations performed using the parallel tempering technique \cite{Hukushima2013}. In this section we present the results only for $\beta=0.6$ and $\beta=0.9$, the summary of all the results is presented in \ref{sec:results} in Tables~\ref{tab:ea-F} to ~\ref{tab:ea-EM-II}.

\begin{figure}
    \centering
    \includegraphics[width=0.95\linewidth]{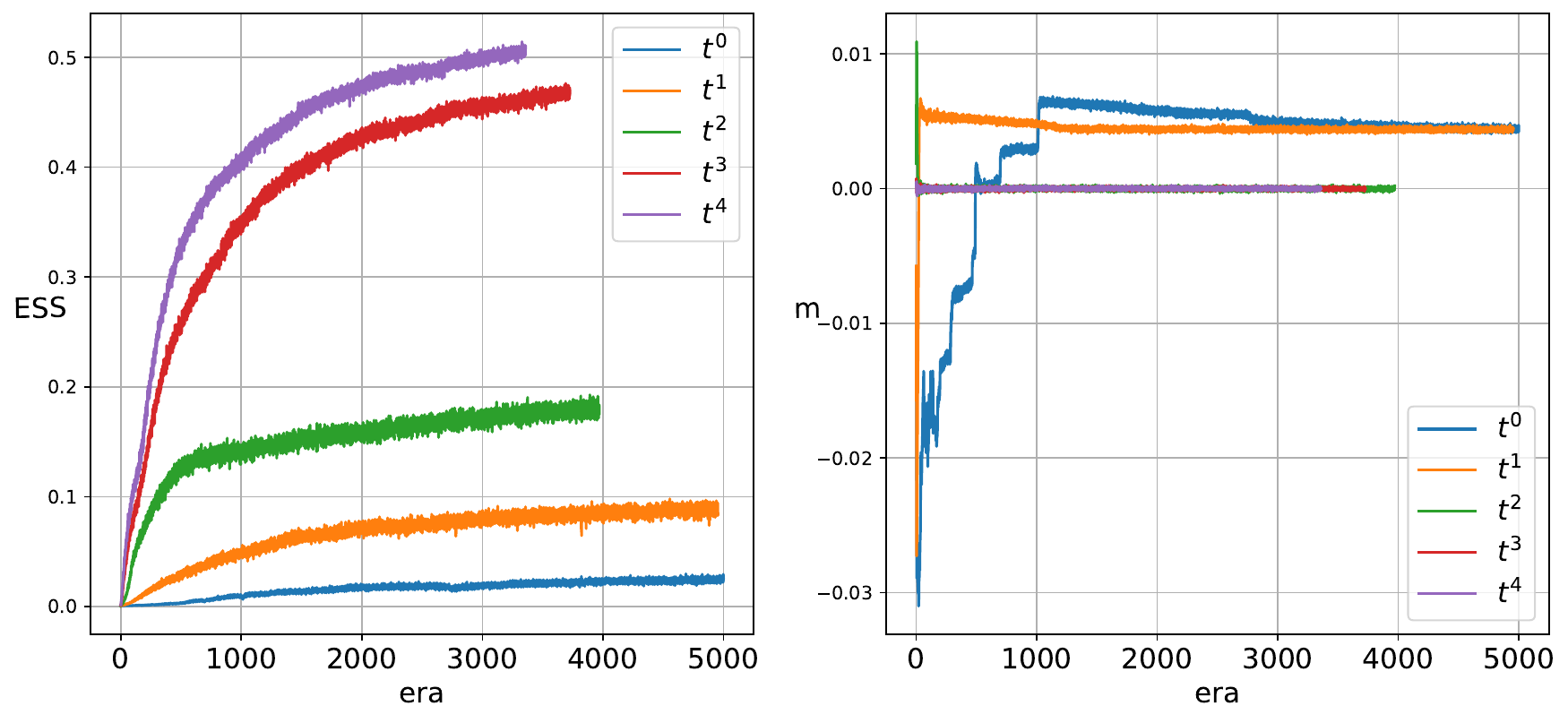}
    \caption{Training history for the Edwards-Anderson model on $32\time 32$ lattice at $\beta=0.6$. ESS (left) and $m$ (right).  The horizontal axis marks the eras where an era is a 100 weights update using the batch of $4096$ samples. For clarity we present a moving average over $100$ weights updates.}
    \label{fig:ea-ess-m-b06000}
\end{figure}

\begin{figure}
    \centering
    \includegraphics[width=0.5\linewidth]{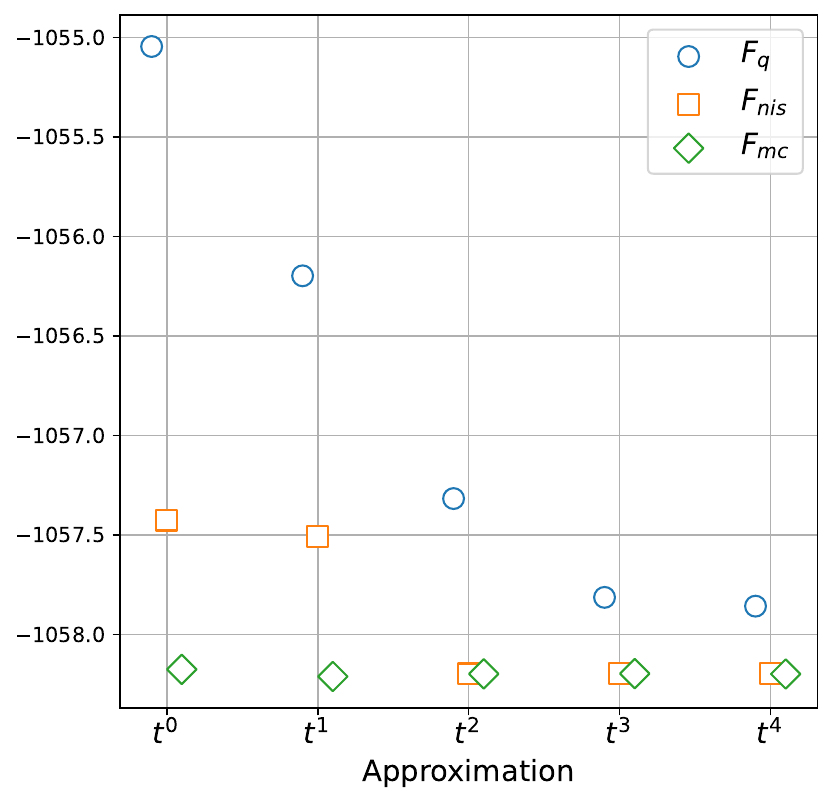}\includegraphics[width=0.5\linewidth]{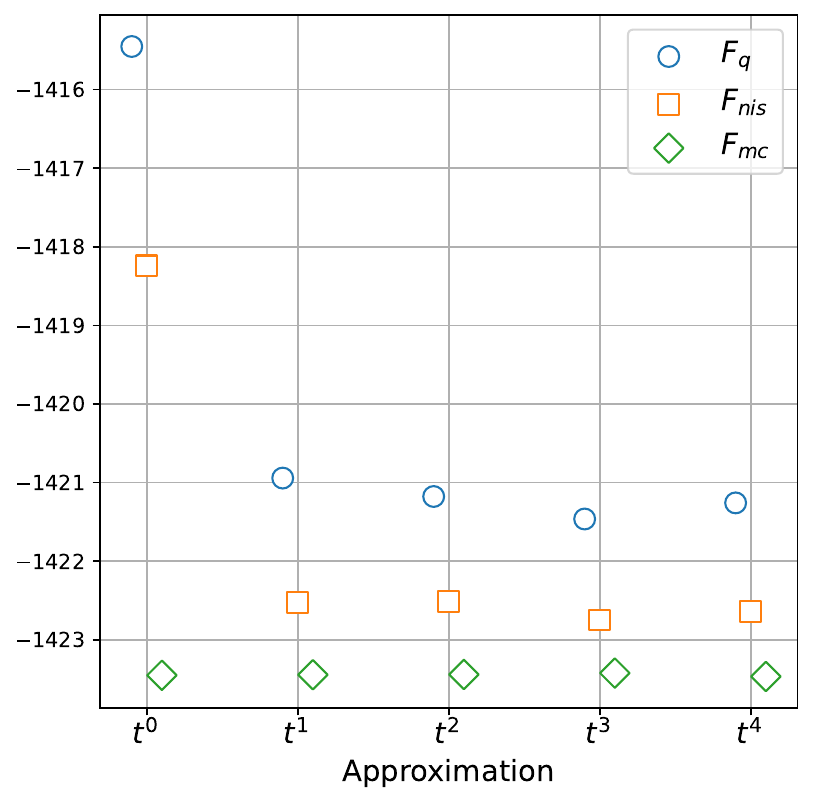}
    \caption{Different estimates of $F$ for the Edwards-Anderson model at $\beta=0.6$ (left) and $\beta=0.9$ (right).}
    \label{fig:ea-F-6}
\end{figure}

Looking at Figure~\ref{fig:ea-ess-m-b06000}, where the training history for the $\beta=0.6$ is presented, we observe similar behavior as for the Ising model.
We see a big increase in the efficiency of training from $t^2$ to $t^3$. Looking at the magnetization in the same Figure (right) and estimates of $F$ in the Figure~\ref{fig:ea-F-6} (left), we notice that the model is able to train  starting with the $t^2$ approximation, albeit with significantly lower ESS  (see also Tables~\ref{tab:ea-F} to \ref{tab:ea-EM-II}). 

\begin{figure}
    \centering
    \includegraphics[width=0.95\linewidth]{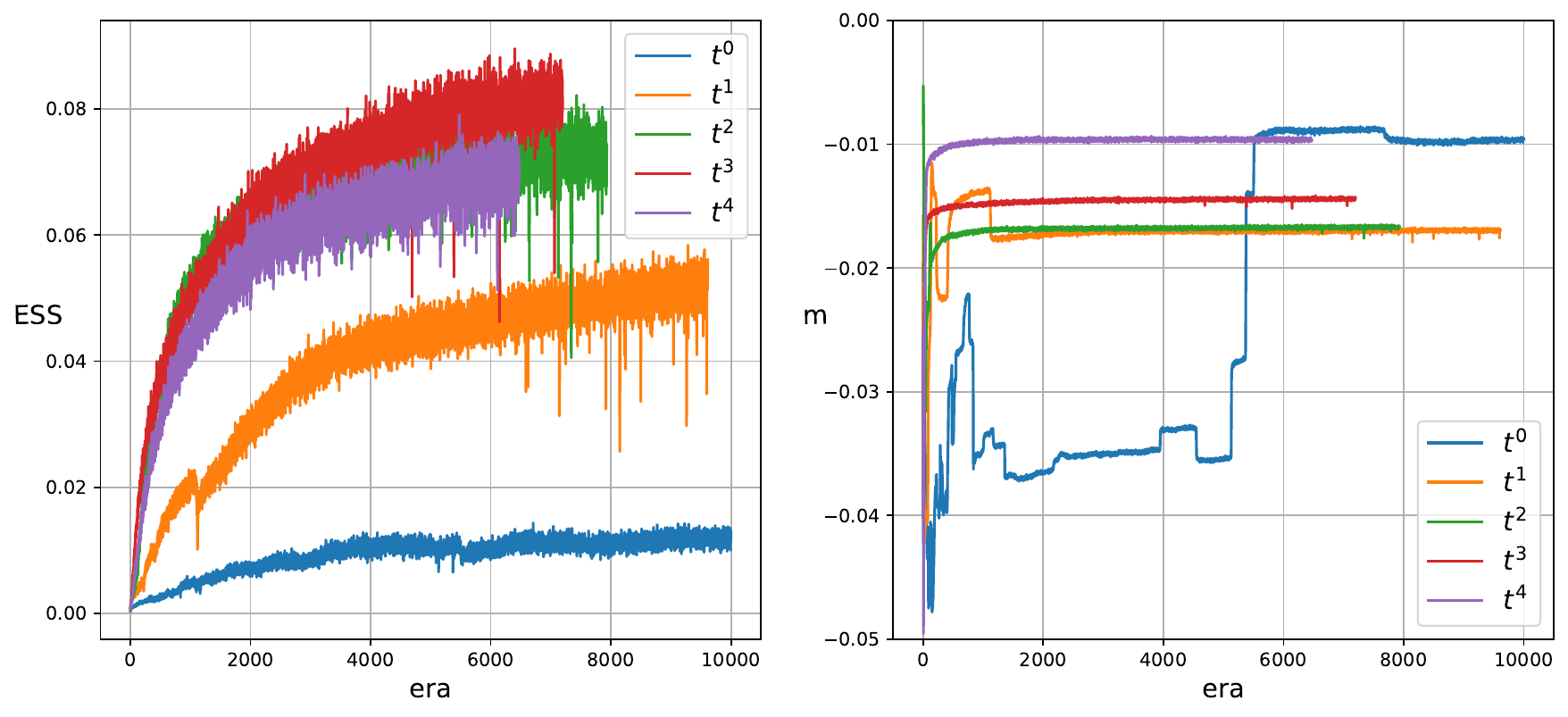}
    \caption{Results for the Edwards-Anderson model on $32\time 32$ lattice at $\beta=0.9$. ESS (left) and $m$ (right).  The horizontal axis marks the eras where an era is a 100 weights update using the batch of $4096$ samples. For clarity we present a moving average over $200$ weights updates.}
    \label{fig:ea-ess-m-b09000}
\end{figure}

The situation is different for $\beta=0.9$, as presented in Figure~\ref{fig:ea-ess-m-b09000}.
While we still see a  fourfold increase in $ESS$ from $t^1$ to $t^3$, it is still at the level of only $5\%$. Looking at Figure~\ref{fig:ea-F-6} (right) and into Tables~\ref{tab:ea-F}
 and \ref{tab:ea-EM-II}, we can see that none of the models trained properly, each exhibiting a mode collapse as indicated by the value of parameter $\bar{w}$. This mode collapse seems to be associated with broken $Z_2$ symmetry, as we have a small but non-zero average magnetization. We should also mention that the best results are obtained for $t^3$,  which is in agreement with the results in Table~\ref{tab:ea-F-q}.

\section{Conclusions and outlook}

In this contribution, we have advocated for the use of probability priors to enhance the autoregressive neural generators for spin systems. The idea is to approximate the conditional probabilities by using only a subset of neighboring spins. This can be done in a systematic way using a weak expansion in $\tanh\beta$. We have derived formulas up to order $t^4$ for the general case of nearest neighbor interactions. We have tested our approach on the two-dimensional Ising model and $J=\pm 1$ Edwards-Anderson spin glass. In both cases, the results are very encouraging, the addition of priors significantly improves the quality of training. While the references \cite{Biazzo2023} and \cite{Biazzo2024} also introduce the priors, albeit using a different architecture, they only use the nearest neighbors in the approximation. Our results prove that including further spins can significantly improve the training, sometimes making the difference between an architecture that can be trained and one that cannot. 

We have derived the approximate probabilities using an expansion in powers of $\tanh\beta$ and this can be done for any nearest neighbors model. However, those approximations do not have to be derived in such a systematic way. Any approximation that is closer to the target distribution than the "blank slate" approach can potentially be beneficial. The search for such approximations for different physical models is the subject of ongoing work. 

We consider this paper as a proof of concept, and to this end, we have used only a very simple neural network architecture. We have also omitted by design any possible enhancements, like incorporating the $Z_2$ symmetry or $\beta$ annealing,  which have been proven to increase the quality of the training. All those things are orthogonal to the approach proposed here and can be used jointly. In particular, the priors can be incorporated into any autoregressive architecture. We have already incorporated the priors into a transformer based architecture with similarly promising results \cite{Transformers}.

\section*{Acknowledgments}
We gratefully acknowledge Polish high-performance computing infrastructure PLGrid (HPC Center: ACK Cyfronet AGH) for providing computer facilities and support within computational grant no. PLG/2025/018811.
T.S. and D.Z. kindly acknowledge the support of the Polish National Science Center (NCN) Grant No.\,2021/43/D/ST2/03375.  P.K. acknowledges the support of the Polish National Science Center (NCN) grant No. 2022/46/E/ST2/00346.

\bibliography{neumc}

\appendix

\section{Implementation}
\label{sec:implementation}

\begin{figure}
    \centering
    \includegraphics[width=0.95\linewidth]{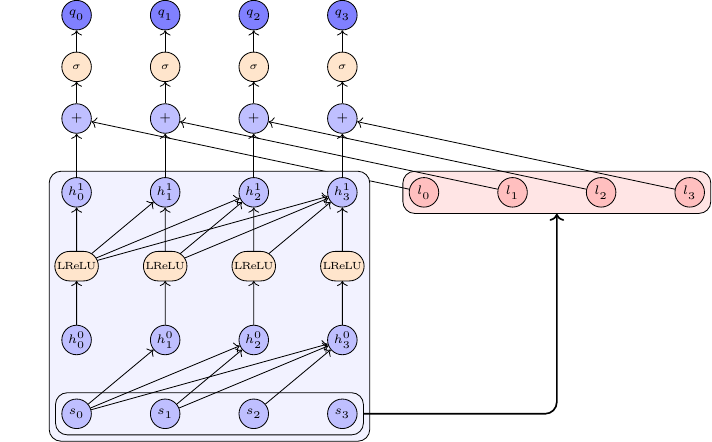}
    \caption{Schema of the model architecture. The neural network block (light blue) can be exchanged for any autoregressive model. The $l_i$ are the approximate logits. }
    \label{fig:implementation}
\end{figure}

As our aim was only to provide the concept of proof for the probability priors, we used the simple architecture, consisting of two dense layers with \texttt{LeakyReLu} activation function in between. The logits of approximate probabilities are added to the output of the last layer and then passed through the logistic function (see Figure~\ref{fig:implementation}).

The most computationally demanding part of the training is sampling the configurations, as we need to calculate each $q(s_i|\sless{i})$ separately in order. The naive approach would be to propagate a whole vector of spins through the network each time (with the first $i$ spins set to already defined values) and  take only the output $i$, disregarding the rest. That, however, would be very wasteful. The better approach is to cache all the previous outputs of each layer and use them to calculate only the output $i$ at each stage. This amounts to effectively multiplying the input by one row of the weight matrix at each step. That greatly reduces the amount of time needed for sampling. When computing the probability of a configuration, which is needed for loss calculations, we just pass the whole input vector through the network at once. 

This optimization implies that we also need two ways of calculating the approximate probabilities: one that does it one at a time and the other all at the same time. For the Ising model, this is relatively easy. The formulas \eqref{eq:ising-logit-1},\eqref{eq:ising-logit-2},\eqref{eq:ising-logit-3} and \eqref{eq:ising-logit-4} can be written in terms of a kernel multiplication (see Figure~\ref{fig:p-cond-27}). So we can use the convolution if we need to calculate all logits at once. Otherwise, we just apply the kernel at one position in the spins lattice. This calculation introduces a $50\%$ overhead (see Table~\ref{tab:timings}).
\begin{table}[]
    \centering
    \begin{tabular}{|c|c|c|c|c|c|}
    \hline\hline
        & $t^0$ & $t^1$ &  $t^2$ &  $t^3$ &  $t^4$  \\\hline\hline 
    Ising    & 1.0 & 1.5& 1.5 & 1.5 & 1.5 \\\hline
     EA    & 1.0 & 1.5 & 1.8 & 2.0 & 2.2\\\hline\hline
    \end{tabular}
    \caption{Relative increase in time needed for one epoch of training (batch size = 4096). }
    \label{tab:timings}
\end{table}

\begin{figure}
    \centering
    \includegraphics[width=0.95\linewidth]{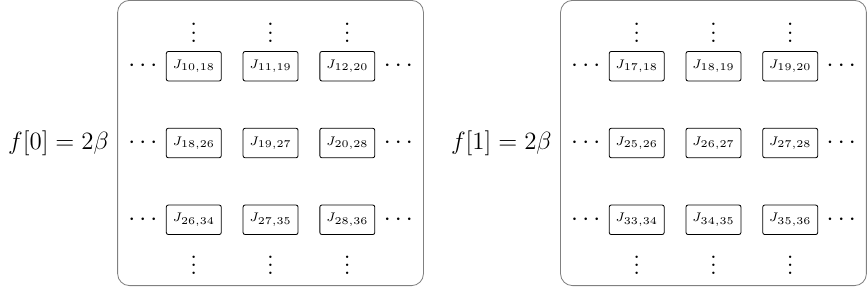}
    \caption{Factors needed for $t^1$ approximation in Edwards-Anderson model.}
    \label{fig:ea-t1}
\end{figure}

\begin{figure}
    \centering
    \includegraphics[width=0.7\linewidth]{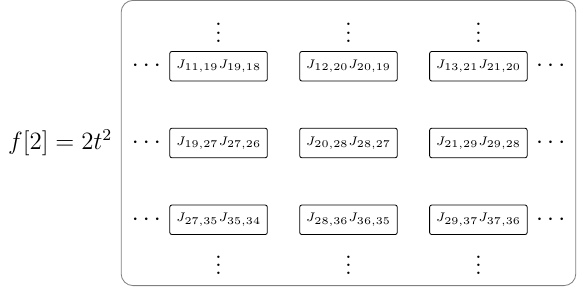}
    \caption{Factors needed for $t^2$ approximation in Edwards-Anderson model.}
    \label{fig:ea-t2}
\end{figure}

In the case of the Edwards-Anderson model, the situation is more involved: we cannot use the convolutions, as the kernel would depend on the position of the spin due to different link variables at each location. We start by precalculating all the products of link variables required in equations \eqref{eq:ea-logit-1} to \eqref{eq:ea-logit-4}. For example, for the $t^2$ approximation, spin $i$ depends on three previous spins and so we need three tables of precalculated factors. Portions of those tables are presented in Figures~\ref{fig:ea-t1} and \ref{fig:ea-t2}. 
With those tables, we can next calculate the required logit by simple multiplication. For example, the logit for the $\s{27}$ spin would be calculated as 
\begin{lstlisting}
f[0,3,3]*s[:,3,5] +f[1,3,3]*s[:,4,4] +f[2,3,3]*s[:,4,6]  
\end{lstlisting}
assuming that \texttt{s}  represents an array of spins as presented in Figure~\ref{fig:p-cond-27}. The first dimension is the batch dimension. All the digits can be calculated at the same time using the following line of code
\begin{lstlisting}
f[0]*s[:,0:8,2:10] +f[1]*s[:,1:9,1:9] +f[2]*s[1:9,3:11]
\end{lstlisting}
Those calculations introduce an overhead  of up to $220\%$ (see Table~\ref{tab:timings}). 

\section{Results}
\label{sec:results}

In this section we present the results for estimators of $F$ and the average energy, magnetization and absolute value of magnetization. Each 
value was obtained using $2^{20}$ samples from neural models and approximately $4\times 10^5$ samples from Monte Carlo. The errors were calculated using the bootstrap method. 

\begin{table}
\begin{center}
\begin{tabular}{|r|S[table-format=1.6]|S[table-format=5.6]|S[table-format=5.6]|S[table-format=5.6]|S[table-format=1.6]|}
\hline\hline
& {ESS} &{$F_q$} & {$F_{nis}$} & {$F_{mc}$} & {$\bar{w}$} \\\hline\hline
\multicolumn{6}{|c|}{$\beta=0.40000$ $F=-900.4783078$} \\\hline
$t^0$ & 0.003(4) & -897.999(2) & -899.85(4) & -900.48(1) & 0.53(3)\\
$t^1$ & 0.152(4) & -899.547(1) & -900.480(2) & -900.476(4) & 1.003(4)\\
$t^2$ & 0.505(2) & -900.1351(8) & -900.4777(9) & -900.475(2) & 1.002(2)\\
$t^3$ & 0.8139(3) & -900.3745(5) & -900.4777(4) & -900.4767(9) & 1.0010(9)\\
$t^4$ & 0.8303(3) & -900.3844(4) & -900.4777(4) & -900.4770(8) & 1.0007(9)\\\hline\hline
\multicolumn{6}{|c|}{$\beta=0.42000$ $F=-924.4135102$} \\\hline
$t^0$ & 0.005(5) & -921.899(2) & -923.66(2) & -924.42(1) & 0.47(1)\\
$t^1$ & 0.298(2) & -923.794(1) & -924.413(2) & -924.414(2) & 0.998(3)\\
$t^2$ & 0.459(2) & -924.0173(8) & -924.414(1) & -924.413(2) & 1.000(2)\\
$t^3$ & 0.6676(7) & -924.2108(7) & -924.4137(7) & -924.414(1) & 1.000(1)\\
$t^4$ & 0.6798(6) & -924.2207(6) & -924.4138(6) & -924.414(1) & 1.000(1)\\\hline\hline
\multicolumn{6}{|c|}{$\beta=0.44069$ $F=-952.6480795$} \\\hline
$t^0$ & 0.004(4) & -950.298(2) & -951.88(3) & -952.65(1) & 0.46(2)\\
$t^1$ & 0.014(8) & -950.877(1) & -951.954(9) & -952.644(6) & 0.501(5)\\
$t^2$ & 0.31(1) & -952.155(1) & -952.647(2) & -952.649(2) & 0.998(3)\\
$t^3$ & 0.35(1) & -952.2659(8) & -952.647(1) & -952.648(2) & 0.999(2)\\
$t^4$ & 0.381(6) & -952.2667(9) & -952.647(1) & -952.647(2) & 1.000(2)\\\hline\hline
\multicolumn{6}{|c|}{$\beta=0.50000$ $F=-1051.1049876$} \\\hline
$t^0$ & 0.032(4) & -1048.722(2) & -1050.381(5) & -1051.12(1) & 0.479(5)\\
$t^1$ & 0.13(6) & -1049.273(2) & -1050.418(3) & -1051.103(7) & 0.504(4)\\
$t^2$ & 0.31(1) & -1049.334(2) & -1050.413(1) & -1051.109(6) & 0.499(3)\\
$t^3$ & 0.32(2) & -1049.340(2) & -1050.412(1) & -1051.115(7) & 0.495(4)\\
$t^4$ & 0.770(8) & -1050.9674(4) & -1051.0586(6) & -1051.1042(8) & 0.955(1)\\\hline\hline
\end{tabular}
\end{center}
\caption{\label{tab:ising-F}Different estimates of $F$ for the Ising model. The value beside the $\beta$ is the true value of free energy.}
\end{table}

\begin{table}[]
\begin{center}
\begin{tabular}{|r|c|S[table-format=5.5]|S[table-format=5.5]|S[table-format=5.5]|}
\hline\hline
 & &  {variational} & {NIS} & {MC} \\\hline
\multicolumn{5}{|c|}{$\beta=0.40$} \\\hline
none & $E$ & -1142.12(8) & -1134.7(8) & -1133.9(2)\\
 & $M$ & -146.2(2) & -94(2) & 0.3(4)\\
 & $|M|$ & 228.5(1) & 205(2) & 206.2(2)\\
\hline
t1 & $E$ & -1129.71(7) & -1133.8(1) & -1133.9(2)\\
 & $M$ & 4.8(3) & 0.2(6) & 0.3(4)\\
 & $|M|$ & 206.9(1) & 206.2(3) & 206.2(2)\\
\hline
t2 & $E$ & -1129.28(8) & -1133.81(9) & -1133.9(2)\\
 & $M$ & -1.8(2) & -0.5(4) & 0.3(4)\\
 & $|M|$ & 201.7(1) & 206.0(2) & 206.2(2)\\
\hline
t3 & $E$ & -1131.83(6) & -1134.00(9) & -1133.9(2)\\
 & $M$ & -1.3(3) & 0.0(3) & 0.3(4)\\
 & $|M|$ & 202.9(1) & 206.3(2) & 206.2(2)\\
\hline
t4 & $E$ & -1131.76(8) & -1133.92(8) & -1133.9(2)\\
 & $M$ & 0.8(3) & 0.0(3) & 0.3(4)\\
 & $|M|$ & 203.4(1) & 205.9(2) & 206.2(2)\\
\hline\hline
\multicolumn{5}{|c|}{$\beta=0.42$} \\\hline
none & $E$ & -1295.2(1) & -1271.9(9) & -1267.9(2)\\
 & $M$ & -411.5(3) & -302(4) & 1.4(7)\\
 & $|M|$ & 437.2(2) & 368(3) & 359.9(3)\\
\hline
t1 & $E$ & -1262.4(1) & -1267.9(1) & -1267.9(2)\\
 & $M$ & 5.8(4) & -0.7(8) & 1.4(7)\\
 & $|M|$ & 346.2(2) & 359.5(3) & 359.9(3)\\
\hline
t2 & $E$ & -1260.2(1) & -1268.2(1) & -1267.9(2)\\
 & $M$ & -3.2(4) & 1.0(5) & 1.4(7)\\
 & $|M|$ & 345.1(2) & 360.1(3) & 359.9(3)\\
\hline
t3 & $E$ & -1260.89(9) & -1268.03(9) & -1267.9(2)\\
 & $M$ & 0.9(4) & -0.4(5) & 1.4(7)\\
 & $|M|$ & 341.2(2) & 359.7(2) & 359.9(3)\\
\hline
t4 & $E$ & -1260.6(1) & -1267.97(8) & -1267.9(2)\\
 & $M$ & -0.1(3) & 0.4(5) & 1.4(7)\\
 & $|M|$ & 339.9(2) & 359.8(2) & 359.9(3)\\
\hline\hline
 \end{tabular}
 \end{center}
 \caption{\label{tab:ising-EM-I}Ising model. Part I. }
 \end{table}

 \begin{table}[]
\begin{center}
\begin{tabular}{|r|c|S[table-format=5.5]|S[table-format=5.5]|S[table-format=5.5]|}
\hline\hline
 & &  {variational} & {NIS} & {MC} \\\hline
\multicolumn{5}{|c|}{$\beta=0.44$} \\\hline
none & $E$ & -1494.26(9) & -1484(1) & -1468.2(1)\\
 & $M$ & -739.28(9) & -705(1) & 1(1)\\
 & $|M|$ & 739.26(9) & 703(1) & 670.3(3)\\
\hline
t1 & $E$ & -1479.0(1) & -1468.9(9) & -1468.2(1)\\
 & $M$ & 713.5(1) & 673(3) & 1(1)\\
 & $|M|$ & 713.7(1) & 674(3) & 670.3(3)\\
\hline
t2 & $E$ & -1468.5(1) & -1467.9(2) & -1468.2(1)\\
 & $M$ & 3.8(8) & 1(1) & 1(1)\\
 & $|M|$ & 674.3(2) & 669.6(5) & 670.3(3)\\
t3 & $E$ & -1477.8(1) & -1468.2(2) & -1468.2(1)\\
 & $M$ & -0.5(8) & 2.0(8) & 1(1)\\
 & $|M|$ & 693.8(2) & 670.4(5) & 670.3(3)\\
\hline
t4 & $E$ & -1472.7(1) & -1468.1(1) & -1468.2(1)\\
 & $M$ & -6.1(7) & -1(1) & 1(1)\\
 & $|M|$ & 685.4(2) & 670.1(4) & 670.3(3)\\
\hline\hline
\multicolumn{5}{|c|}{$\beta=0.50$} \\\hline
none & $E$ & -1795.07(5) & -1783(5) & -1787.41(9)\\
 & $M$ & -939.07(2) & -923(1) & 0(2)\\
 & $|M|$ & 939.07(2) & 924(9) & 933.15(4)\\
\hline
t1 & $E$ & -1784.78(5) & -1787.6(2) & -1787.41(9)\\
 & $M$ & -934.24(2) & -933.2(2) & 0(2)\\
 & $|M|$ & 934.24(2) & 933.2(2) & 933.15(4)\\
\hline
t2 & $E$ & -1781.68(5) & -1787.4(1) & -1787.41(9)\\
 & $M$ & 932.07(2) & 933.25(8) & 0(2)\\
 & $|M|$ & 932.07(2) & 933.26(8) & 933.15(4)\\
\hline
t3 & $E$ & -1785.59(5) & -1787.6(1) & -1787.41(9)\\
 & $M$ & 931.29(3) & 933.2(1) & 0(2)\\
 & $|M|$ & 931.29(3) & 933.2(1) & 933.15(4)\\
\hline
t4 & $E$ & -1791.21(6) & -1788.12(6) & -1787.41(9)\\
 & $M$ & -0(1) & -0(1) & 0(2)\\
 & $|M|$ & 935.39(2) & 933.58(5) & 933.15(4)\\
\hline\hline
 \end{tabular}
 \end{center}
 \caption{\label{tab:ising-EM-II}Ising model. Part II}
 \end{table}

\begin{table}
\begin{center}
\begin{tabular}{|r|S[table-format=1.7]|S[table-format=5.6]|S[table-format=5.6]|S[table-format=5.6]|S[table-format=2.6]|}
\hline\hline
& {ESS} &{$F_q$} & {$F_{nis}$} & {$F_{mc}$} & {$\bar{w}$} \\\hline\hline
\multicolumn{6}{|c|}{$\beta=0.30000$} \\\hline
$t^0$ & 0.006(1) & -798.082(2) & -800.86(1) & -800.86(3) & 1.00(3)\\
$t^1$ & 0.5686(9) & -800.5702(7) & -800.8526(9) & -800.851(2) & 1.001(2)\\
$t^2$ & 0.9165(1) & -800.8074(3) & -800.8509(3) & -800.8515(6) & 0.9994(6)\\
$t^3$ & 0.95606(6) & -800.8290(2) & -800.8515(2) & -800.8515(4) & 1.0000(4)\\
$t^4$ & 0.96558(5) & -800.8337(2) & -800.8512(2) & -800.8511(4) & 1.0001(4)\\\hline\hline
\multicolumn{6}{|c|}{$\beta=0.60000$} \\\hline
$t^0$ & 0.006(3) & -1055.045(2) & -1057.43(2) & -1058.18(7) & 0.47(3)\\
$t^1$ & 0.065(7) & -1056.198(2) & -1057.508(4) & -1058.21(1) & 0.495(5)\\
$t^2$ & 0.173(3) & -1057.317(1) & -1058.198(2) & -1058.198(5) & 1.000(5)\\
$t^3$ & 0.463(2) & -1057.8139(7) & -1058.198(1) & -1058.197(2) & 1.001(2)\\
$t^4$ & 0.506(1) & -1057.8580(9) & -1058.1972(9) & -1058.199(2) & 0.998(2)\\\hline\hline
\multicolumn{6}{|c|}{$\beta=0.90000$} \\\hline
$t^0$ & 0.0004(2) & -1415.451(2) & -1418.24(5) & -1423.5(2) & 0.006(1)\\
$t^1$ & 0.012(9) & -1420.943(2) & -1422.53(1) & -1423.45(2) & 0.400(8)\\
$t^2$ & 0.048(4) & -1421.178(2) & -1422.517(5) & -1423.44(1) & 0.396(5)\\
$t^3$ & 0.052(7) & -1421.464(2) & -1422.749(4) & -1423.43(3) & 0.51(2)\\
$t^4$ & 0.040(4) & -1421.259(2) & -1422.645(4) & -1423.47(1) & 0.439(6)\\\hline\hline
\end{tabular}
\end{center}
\caption{\label{tab:ea-F}Different estimates of $F$ for Edwards-Anderson model.}
\end{table}

\begin{table}[]
\begin{center}
\begin{tabular}{|r|c|S[table-format=5.5]|S[table-format=5.5]|S[table-format=5.5]|}
\hline\hline
 & &  {variational} & {NIS} & {MC} \\\hline\hline
\multicolumn{5}{|c|}{$\beta=0.30$} \\\hline
none & $E$ & -592.83(5) & -599.8(5) & -599.64(5)\\
 & $M$ & -0.67(4) & 0.2(5) & 0.00(3)\\
 & $|M|$ & 25.00(2) & 25.3(2) & 25.00(2)\\
\hline
t1 & $E$ & -598.58(4) & -599.62(6) & -599.64(5)\\
 & $M$ & 0.08(3) & -0.03(4) & 0.00(3)\\
 & $|M|$ & 24.98(2) & 25.02(2) & 25.00(2)\\
\hline
t2 & $E$ & -599.37(4) & -599.66(4) & -599.64(5)\\
 & $M$ & -0.10(3) & -0.05(3) & 0.00(3)\\
 & $|M|$ & 24.99(2) & 24.98(2) & 25.00(2)\\
\hline
t3 & $E$ & -599.60(4) & -599.64(5) & -599.64(5)\\
 & $M$ & -0.08(3) & 0.03(3) & 0.00(3)\\
 & $|M|$ & 24.99(2) & 24.98(2) & 25.00(2)\\
\hline
t4 & $E$ & -599.80(4) & -599.74(4) & -599.64(5)\\
 & $M$ & 0.06(3) & 0.05(3) & 0.00(3)\\
 & $|M|$ & 24.99(2) & 24.99(2) & 25.00(2)\\
\hline\hline
\multicolumn{5}{|c|}{$\beta=0.60$} \\\hline
none & $E$ & -1080.63(3) & -1080.4(6) & -1080.78(4)\\
 & $M$ & 3.23(3) & 4.7(3) & -0.04(4)\\
 & $|M|$ & 25.00(2) & 25.0(3) & 25.19(2)\\
\hline
t1 & $E$ & -1079.02(3) & -1080.6(1) & -1080.78(4)\\
 & $M$ & 4.45(3) & 4.3(1) & -0.04(4)\\
 & $|M|$ & 25.08(2) & 25.04(6) & 25.19(2)\\
\hline
t2 & $E$ & -1078.66(3) & -1080.75(8) & -1080.78(4)\\
 & $M$ & -0.15(3) & -0.11(7) & -0.04(4)\\
 & $|M|$ & 25.20(2) & 25.24(4) & 25.19(2)\\
\hline
t3 & $E$ & -1079.74(3) & -1080.78(5) & -1080.78(4)\\
 & $M$ & -0.53(3) & -0.01(5) & -0.04(4)\\
 & $|M|$ & 25.16(2) & 25.15(3) & 25.19(2)\\
\hline
t4 & $E$ & -1079.94(4) & -1080.70(5) & -1080.78(4)\\
 & $M$ & -0.45(3) & 0.00(4) & -0.04(4)\\
 & $|M|$ & 25.17(2) & 25.24(3) & 25.19(2)\\
\hline\hline
\end{tabular}    
\end{center}
    \caption{Edwards-Anderson model. Part I. }
    \label{tab:ea-EM-I}
\end{table}

\begin{table}[]
\begin{center}
\begin{tabular}{|r|c|S[table-format=5.5]|S[table-format=5.5]|S[table-format=5.5]|}
\hline\hline
 & &  {variational} & {NIS} & {MC} \\\hline\hline
\multicolumn{5}{|c|}{$\beta=0.90$} \\\hline
none & $E$ & -1319.91(2) & -1322(3) & -1316.86(3)\\
 & $M$ & -9.60(3) & -9(2) & 0.07(6)\\
 & $|M|$ & 24.36(2) & 23(3) & 28.16(3)\\
\hline
t1 & $E$ & -1317.72(2) & -1317.6(1) & -1316.86(3)\\
 & $M$ & -17.13(3) & -16.4(2) & 0.07(6)\\
 & $|M|$ & 28.56(2) & 28.5(1) & 28.16(3)\\
\hline
t2 & $E$ & -1317.03(2) & -1317.4(1) & -1316.86(3)\\
 & $M$ & -17.26(3) & -16.4(3) & 0.07(6)\\
 & $|M|$ & 28.98(2) & 29.0(1) & 28.16(3)\\
\hline
t3 & $E$ & -1316.32(2) & -1316.2(5) & -1316.86(3)\\
 & $M$ & -15.27(3) & -14.6(3) & 0.07(6)\\
 & $|M|$ & 28.48(2) & 28.2(2) & 28.16(3)\\
\hline
t4 & $E$ & -1317.08(2) & -1317.5(1) & -1316.86(3)\\
 & $M$ & -9.39(4) & -8.5(1) & 0.07(6)\\
 & $|M|$ & 28.31(2) & 28.1(1) & 28.16(3)\\
\hline\hline
\end{tabular}    
\end{center}
    \caption{Edwards-Anderson model. Part II. }
    \label{tab:ea-EM-II}
\end{table}

\end{document}